\documentclass[10pt,twocolumn,letterpaper]{article}
\pdfoutput=1
\usepackage[pagebackref=true,breaklinks=true,letterpaper=true,colorlinks,bookmarks=false,citecolor=citecolor, linkcolor=linkcolor]{hyperref}

\usepackage{iccv}
\usepackage{times}
\usepackage{epsfig}
\usepackage{graphicx}
\usepackage{amsmath}
\usepackage{amssymb}
\usepackage[utf8]{inputenc} 
\usepackage[T1]{fontenc}    
\usepackage{url}            
\usepackage{booktabs}       
\usepackage{amsfonts}       
\usepackage{nicefrac}       
\usepackage{microtype}      
\usepackage{array,colortbl}
\usepackage{xcolor}
\usepackage[capitalise]{cleveref}
\usepackage{caption}
\usepackage{multirow}

\newcommand{\bI}{\mathbf{I}}
\newcommand{\bR}{\mathbf{R}}
\newcommand{\bzero}{\mathbf{0}}
\newcommand{\E}{\mathbb{E}}
\newcommand{\Eb}[2]{\E_{#1}\!\left[#2\right]}
\newcommand{\bx}{\mathbf{x}}
\newcommand{\by}{\mathbf{y}}
\newcommand{\bepsilon}{{\boldsymbol{\epsilon}}}
\newcommand{\bSigma}{{\boldsymbol{\Sigma}}}
\newcommand{\diff}{\mathrm{d}}
\newcommand{\bmu}{{\boldsymbol{\mu}}}
\newcommand{\authorskip}{\hspace{2.5mm}}

\definecolor{citecolor}{HTML}{0071BC}
\definecolor{linkcolor}{HTML}{ED1C24}

\iccvfinalcopy 

\def\iccvPaperID{****} 

\ificcvfinal\fi

\begin{document}

\title{\vspace{-1mm}\Large DIRE for Diffusion-Generated Image Detection\vspace{-3mm}}

\author{Zhendong Wang{\small $^{1}$}$^*$ \authorskip Jianmin Bao{\small$^{2}$}\thanks{Equal contribution.} \authorskip Wengang Zhou{\small$^{1,3}$} \\ 
Weilun Wang{\small$^{1}$} \authorskip Hezhen Hu{\small$^{1}$} \authorskip Hong Chen{\small$^{4}$} \authorskip Houqiang Li{\small$^{1,3}$} \\
\small $^{1}$ CAS Key Laboratory of GIPAS, EEIS Department, University of Science and Technology of China \\
\small
$^{2}$ Microsoft Research Asia \\
\small
$^{3}$ Institute of Artificial Intelligence, Hefei Comprehensive National Science Center\\
\small
$^{4}$ Merchants Union Consumer Finance Company\\
\small
{\tt\small \{zhendongwang,wwlustc,alexhu\}@mail.ustc.edu.cn} \\
{\tt\small jianbao@microsoft.com, \{zhwg,lihq\}@ustc.edu.cn, chenhong@mucfc.com}
}

\maketitle
\ificcvfinal\thispagestyle{empty}\fi

\begin{abstract}
    \vspace{-3mm}
   Diffusion models have shown remarkable success in visual synthesis, but have also raised concerns about potential abuse for malicious purposes. In this paper, we seek to build a detector for telling apart real images from diffusion-generated images. We find that existing detectors struggle to detect images generated by diffusion models, even if we include generated images from a specific diffusion model in their training data. To address this issue, we propose a novel image representation called \textbf{DI}ffusion \textbf{R}econstruction \textbf{E}rror~(\textbf{DIRE}), which measures the error between an input image and its reconstruction counterpart by a pre-trained diffusion model. We observe that diffusion-generated images can be approximately reconstructed by a diffusion model while real images cannot. 
   It provides a hint that DIRE can serve as a bridge to distinguish generated and real images. 
   DIRE provides an effective way to detect images generated by most diffusion models, and it is general for detecting generated images from unseen diffusion models and robust to various perturbations. Furthermore, we establish a comprehensive diffusion-generated benchmark including images generated by eight diffusion models to evaluate the performance of diffusion-generated image detectors. 
   Extensive experiments on our collected benchmark demonstrate that DIRE exhibits superiority over  previous generated-image detectors.
   The code and dataset are available at \url{https://github.com/ZhendongWang6/DIRE}.
\end{abstract}

\vspace{-5mm}
\section{Introduction}

Recently, Denoising Diffusion Probabilistic Models~(DDPMs)~\cite{DDPM,sohl2015deep} have set up a new paradigm in image generation due to their strong ability to generate high-quality images. There arises plenty of studies~\cite{iDDPM,ADM,DDIM,PNDM,LDM} exploring the improvement of the network architecture, acceleration of sampling, and so on. As users enjoy the strong generation capability of diffusion models, there are concerns about potential privacy problems. For example, diffusion models may memorize individual images from their training data and emit them at the generation stage~\cite{carlini2023extracting, zhu2023data}. Moreover, some attackers may develop new deepfake techniques based on diffusion models. 
Therefore, it is an urgent demand for a diffusion-generated image detector.

\begin{figure}[t] 
    \centering 
    \includegraphics[width=\linewidth]{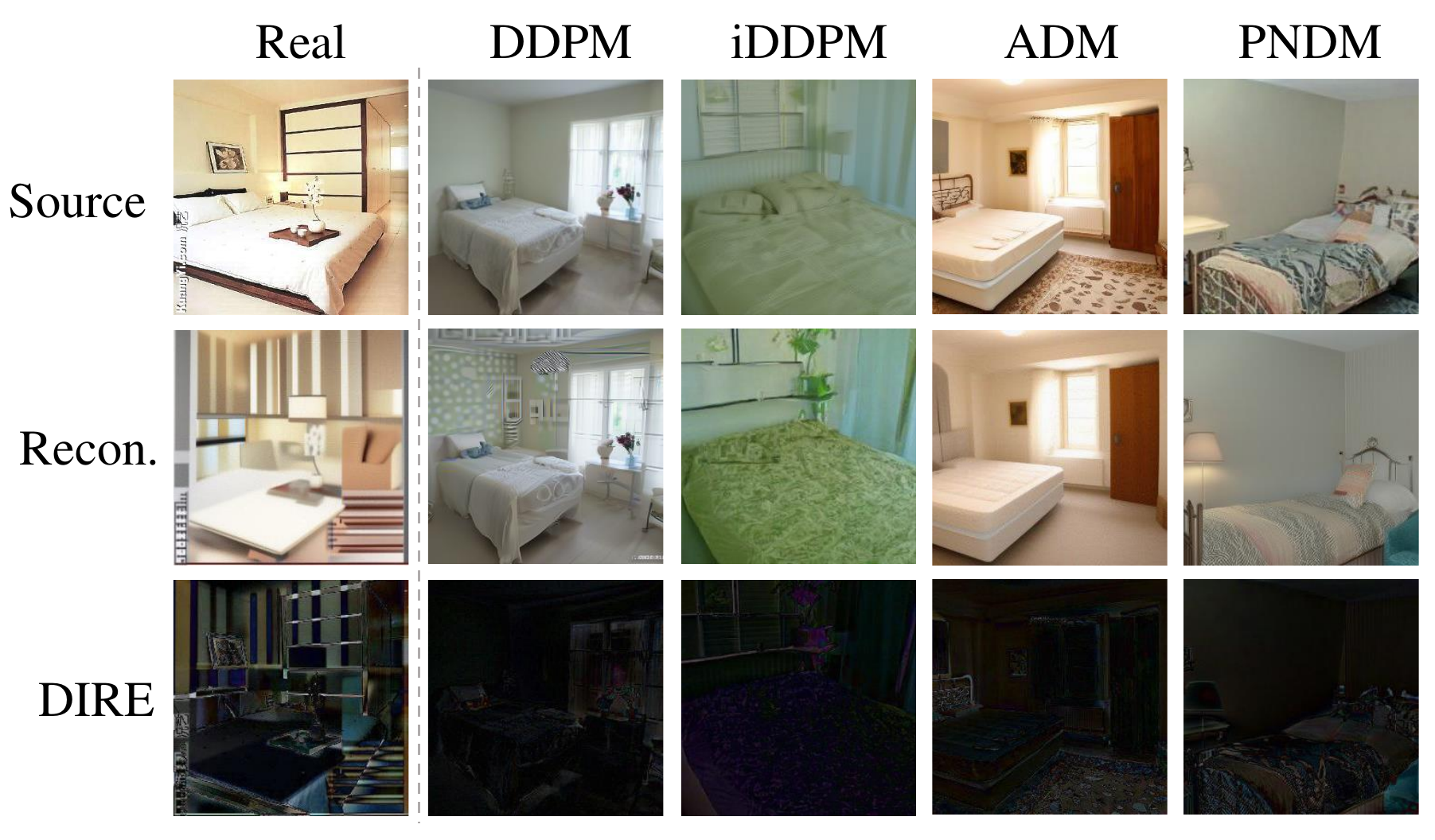}
    \vspace{-2em}
    \caption{\textbf{The DIRE representation} of a real image and four generated images from diffusion models: DDPM~\cite{DDPM}, iDDPM~\cite{iDDPM}, ADM~\cite{ADM}, and PNDM~\cite{PNDM}, respectively. The DIREs of real images tend to have larger values compared to diffusion-generated images.}
    \label{fig:teaser}
    \vspace{-1em}
\end{figure}

Our focus in this work is to develop a general diffusion-generated image detector. We notice that there are various detectors for detecting generated images available. Despite the fact that most diffusion models employ CNNs as the network, the generation processes between diffusion models and previous generators~(\eg, GAN, VAE) are entirely different, rendering previously generated image detectors ineffective. A na\"ive thought is to train a CNN binary classifier on diffusion-generated and real images. However, we find that such a na\"ive scheme suffers limited generalization to unseen diffusion-generated images.

\begin{figure}[t] 
    \centering 
    \includegraphics[width=1.0\linewidth]{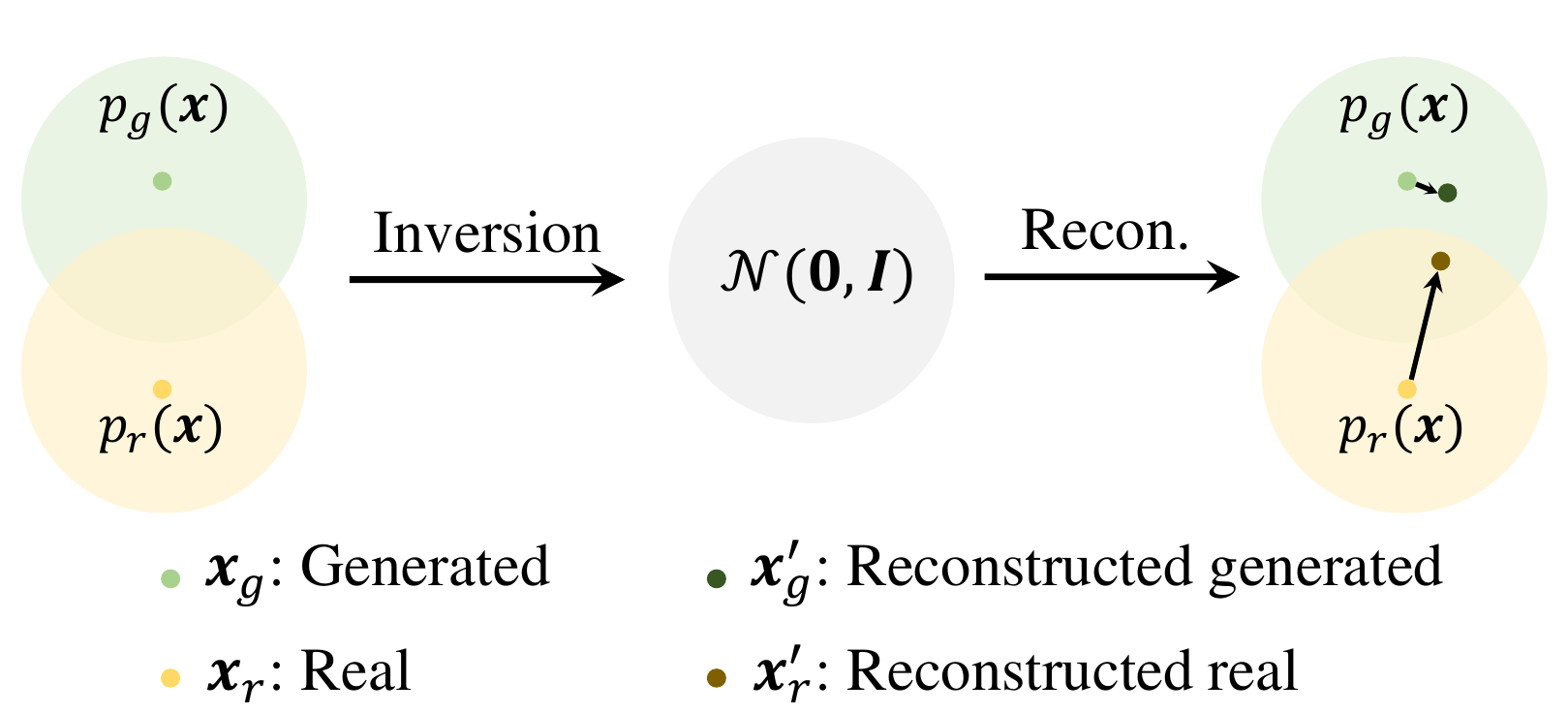}
    \vspace{-2em}
    \caption{\textbf{Illustration of the difference between a real sample and a generated sample from the DIRE perspective.} $p_g(\bx)$ represents the distribution of generated images while $p_{r}(x)$ represents the distribution of real images. $\bx_g$ and $\bx_r$ represent a generated sample and a real sample, respectively. Using the inversion and reconstruction process of DDIM~\cite{DDIM}, $\bx_g$ and $\bx_r$ become $\bx'_g$ and $\bx'_r$, respectively. After the reconstruction, $\bx'_r$ is actually within the $p_g(\bx)$, which leads to a noticeably different DIRE in real samples compared to generated samples.}
    \label{fig:distribution}
    \vspace{-1em}
\end{figure}

In this paper, we propose a novel image representation, called \textbf{DI}ffusion \textbf{RE}construction \textbf{E}rror~(\textbf{DIRE}), for detecting diffusion-generated images. The hypothesis behind DIRE is that images produced by diffusion processes can be reconstructed more accurately by a pre-trained diffusion model compared to real images. The diffusion reconstruction process involves two steps: (1) inverting the input image $\bx$ and mapping it to a noise vector $\bx_T$ in the noise space $\mathcal{N}(\bzero, \bI)$, and (2) reconstructing the image $\bx'$ from $\bx_T$ using a denoising process. The DIRE is calculated as the difference between $\bx$ and $\bx'$. As a sample $\bx_g$ from the generated distribution $p_g(\bx)$ and its reconstruction $\bx'_g$ belong to the same distribution, the DIRE value for $\bx_g$ would be relatively low. Conversely, the reconstruction of a real image $\bx_r$ is likely to differ significantly from itself, resulting in a high amplitude in DIRE. This concept is depicted in Figure~\ref{fig:distribution}.

DIRE offers a reliable method for differentiating between real and diffusion-generated images. By training a simple binary classifier based on DIRE, it becomes possible to detect diffusion-generated images with ease. The DIRE is general and flexible since it can generalize to images generated by unseen diffusion models during inference time. It only assumes the distinct reconstruction errors of real images and generated ones as shown in Figure~\ref{fig:teaser}.

To evaluate the diffusion-generated image detectors, we create a comprehensive diffusion-generated dataset, the \textit{DiffusionForensics} dataset, including images generated by eight different diffusion models trained on LSUN-Bedroom~\cite{yu2015lsun} and ImageNet~\cite{deng2009imagenet}. DiffusionForensics involves unconditional, conditional, and text-to-image diffusion generation models. We will release the dataset to facilitate a good benchmark for diffusion-generated image detection.

Extensive experiments show that the DIRE representation significantly enhances generalization ability. We show that our framework achieves a remarkably high detection accuracy and average precision on generated images from unseen diffusion models, as well as robustness to various perturbations. In comparison with existing generated image detectors, our framework largely exceeds the competitive state-of-the-art methods.

Our main contributions are three-fold as follows.
\begin{itemize}
    \vspace{-2mm}
    \item We propose a novel image representation called DIRE for detecting diffusion-generated images. 
    \vspace{-2mm}
    \item We set up a new dataset, DiffusionForensics, for benchmarking the diffusion-generated image detectors.
    \vspace{-2mm}
    \item Extensive experiments demonstrate that the proposed DIRE sets a state-of-the-art performance in diffusion-generated detection.
\end{itemize}

\section{Related Work}

Since our focus is to detect diffusion-generated images and the proposed DIRE representation is based on the reconstruction error by a pre-trained diffusion model, we briefly introduce recent diffusion models in image generation and generalizable generated image detection in this section.

\subsection{Diffusion Models for Image Generation} 
Inspired by nonequilibrium thermodynamics~\cite{sohl2015deep}, Ho~\etal~\cite{DDPM} propose a new generation paradigm, denoising diffusion probabilistic models~(DDPMs), which achieves a competitive performance compared to PGGAN~\cite{karras2017progressive} on 256 $\times$ 256 LSUN~\cite{yu2015lsun}. Since then, more and more researchers turn their attention to diffusion models for improving the architectures~\cite{ADM, LDM}, accelerating sampling speed~\cite{iDDPM,DDIM,PNDM,lu2022dpm}, exploring downstream tasks~\cite{hertz2022prompt,nichol2021glide,avrahami2022blended,parmar2023zero}, and \etc. Nichol~\etal~\cite{iDDPM} find that learning variances of the reverse process in DDPMs can contribute to an order of magnitude fewer sampling steps. Song~\etal~\cite{DDIM} generalize DDPMs via a class of non-Markovian diffusion processes into  denoising diffusion implicit models~(DDIMs), which leads to more high-quality samples with fewer sampling steps. A later work ADM~\cite{ADM} finds a much more effective architecture and further achieves a state-of-the-art performance compared to other generative models with classifier guidance. From the perspective that DDPMs can be treated as solving differential equations on manifolds, Liu~\etal~\cite{PNDM} propose pseudo numerical methods for diffusion models~(PNDMs), which further improves sampling efficiency and generation quality. 

Besides unconditional image generation, there are also plenty of text-to-image generation works based on diffusion models~\cite{LDM,vq-diffusion,ramesh2022hierarchical,saharia2022photorealistic,chen2022re,ruiz2022dreambooth}. Among them, VQ-Diffusion~\cite{vq-diffusion} is based on a VQ-VAE~\cite{van2017neural} and models the latent space  by a conditional variant of DDPMs. Another typical work is LDM~\cite{LDM} that conditions the diffusion models on the input by cross-attention mechanism, and proposes latent diffusion models by introducing latent space~\cite{esser2021taming}. Recent popular Stable Diffusion v1 and v2 are based on LDM~\cite{LDM} and further improved to achieve a surprising generation performance.

\subsection{Generalizable Generated Image Detection}
Generated image detection has been widely explored over the past years. Earlier researchers focus on detecting generated images leveraging hand-crafted features, such as color cues~\cite{mccloskey2018detecting}, saturation cues~\cite{mccloskey2019detecting}, blending artifacts~\cite{li2020face}, co-occurrence features~\cite{nataraj2019detecting}. Marra~\etal~\cite{marra2018detection} study several classical deep CNN classifiers~\cite{huang2017densely,szegedy2016rethinking,chollet2017xception} to detect images generated by image-to-image translation networks. However, they do not consider the generalization capability to unseen generation models. In another work, Wang~\etal~\cite{CNNDetection} notice this challenge and claim that training a simple classifier on ProGAN-generated images can generalize to other unseen GAN-based generated images well. However, their strong generalization capability relies on their large-scale training and 20 different models each trained on a different LSUN~\cite{yu2015lsun} object category. 

Besides detection by spatial artifacts, there are also frequency-based methods~\cite{frank2020leveraging,zhang2019detecting}.
Frank~\etal~\cite{frank2020leveraging} present that in the frequency domain, GAN-generated images are more likely to expose severe artifacts mainly caused by upsampling operations in previous GAN architectures. Zhang~\etal~\cite{zhang2019detecting} propose a GAN simulator, AutoGAN, to simulate the artifacts produced by standard GAN pipelines. Then they train a detector on the spectrum input on the synthesized images. It can generalize to unseen generation models to some extent. Marra~\etal~\cite{marra2019gans} and Yu~\etal~\cite{yu2019attributing} suggest detecting generated images by fingerprints that are often produced during GAN generation. A recent work~\cite{mandelli2022detecting} proposes a detector based on an ensemble of EfficientNet-B4~\cite{tan2019efficientnet} to alleviate the generalization problem. 

However, with the boosting development of diffusion models, a general and robust detector for detecting images generated by diffusion models has not been explored. We note that some recent works also notice the diffusion-generated image detection problem~\cite{ricker2022towards,corvi2022detection}. Different from them, the focus of our work is exploring a generalizable detector for wide-range diffusion models.

\section{Method}

In this paper, we present a novel representation named \textbf{DI}ffusion \textbf{R}econstruction \textbf{E}rror ~(\textbf{DIRE}) for diffusion-generated image detection.
DIRE measures the error between an input image and its reconstruction by a pre-trained diffusion model.
We observe that diffusion-generated images can be more approximately reconstructed by a pre-trained diffusion model compared to real images.
Based on this, the DIRE provides discriminative properties for distinguishing diffusion-generated images from real images.
The rest of this section is organized as follows.
We begin with reviewing DDPMs, and the inversion and reconstruction process of the DDIM~\cite{DDIM}.
Then we present details of DIRE for diffusion-generated image detection.
Finally, we introduce a new dataset, \ie, DiffusionForensics, for evaluating diffusion-generated image detectors. 

\subsection{Preliminaries}
\noindent \textbf{Denoising Diffusion Probabilistic Models~(DDPMs).} Diffusion models are first proposed in \cite{sohl2015deep} inspired by non-equilibrium thermodynamics, and achieve strong performance in image generation~\cite{DDPM, iDDPM,ADM, LDM}. They define a Markov chain of diffusion steps that slowly add Gaussian noise to data until degenerating it into isotropic Gaussian distribution~(forward process), and then learn to reverse the diffusion process to generate samples from the noise~(reverse process). The Markov chain in the forward process is defined as:
\begin{equation}
q(\bx_t|\bx_{t-1}) = \mathcal{N}(\bx_t;\sqrt{\frac{\alpha_t}{\alpha_{t-1}}}\bx_{t-1},(1-\frac{\alpha_t}{\alpha_{t-1}}) \bI),\label{eq:forwardprocess}
\end{equation}
in which $\bx_t$ is the noisy image at the $t$-th step and $\alpha_1,\dots, \alpha_T$ is a predefined schedule, with $T$ denotes the total steps. 

An important property brought by the Markov chain is that we can obtain $\bx_t$ from $\bx_0$ directly via:
\begin{equation}
q(\bx_t|\bx_0) = \mathcal{N}(\bx_t;\sqrt{\alpha_t}\bx_0,(1-\alpha_t) \bI). \label{eq:x_t}
\end{equation}
The reverse process in \cite{DDPM} is also defined as a Markov chain:
\begin{equation}
    p_\theta(\bx_{t-1}|\bx_t) = \mathcal{N}(\bx_{t-1}; \bmu_\theta(\bx_t, t), \bSigma_\theta(\bx_t, t)).\label{eq:reverseprocess}
\end{equation}
Diffusion models use a network $p_{\theta}(\bx_{t-1}|\bx_t)$
to fit the real distribution $q(\bx_{t-1}|\bx_t)$. The overall simplified optimization target is a sampling and denoising process as follows,
\begin{equation}
    L_\mathrm{simple}(\theta) = \Eb{t, \bx_0, \bepsilon}{ \left\| \bepsilon - \bepsilon_\theta(\sqrt{\alpha_t} \bx_0 + \sqrt{1-\alpha_t}\bepsilon, t) \right\|^2},\label{eq:training_objective_simple}
\end{equation}
where $\bepsilon \sim \mathcal{N}(\bzero, \bI)$.

\noindent \textbf{Denoising Diffusion Implicit Models~(DDIMs).} 
DDIM~\cite{DDIM} proposes a new deterministic method for accelerating the iterative process without the Markov hypothesis.
The new reverse process in DDIM is as follows,
\begin{equation}
\begin{aligned}
    \bx_{t-1} = \sqrt{\alpha_{t-1}} \left(\frac{\bx_t - \sqrt{1 - \alpha_t} \bepsilon_\theta(\bx_t, t)}{\sqrt{\alpha_t}}\right) \\
    + \sqrt{1 - \alpha_{t-1} - \sigma_t^2} \cdot \bepsilon_\theta(\bx_t, t) +\sigma_t \epsilon_t.\label{eq:ddim_reverse}
\end{aligned}
\end{equation}
If $\sigma_t=0$, the reverse process becomes deterministic~(reconstruction process), in which one noise sample determines one generated image. Furthermore when $T$ is large enough~(\eg, $T=1000$), Eqn.~\eqref{eq:ddim_reverse} can be seen as Euler integration for solving ordinary differential equations (ODEs):
\begin{equation}
    \frac{\bx_{t-\Delta t}}{\sqrt{\alpha_{t-\Delta t}}}  = \frac{\bx_t}{\sqrt{\alpha_t}}  + \left(\sqrt{\frac{1 - \alpha_{t-\Delta t}}{\alpha_{t-\Delta t}}} - \sqrt{\frac{1 - \alpha_{t}}{\alpha_t}}\right) \bepsilon_\theta(\bx_t, t). \label{eq:ddim_euler}
\end{equation}

Suppose $\sigma=\sqrt{1-\alpha}/\sqrt{\alpha}$, $\bar\bx=\bx/\sqrt{\alpha}$, the corresponding ODE becomes:
\vspace{-2mm}
\begin{equation}
    \diff \bar{\bx}(t) = \bepsilon_\theta\left(\frac{\bar{\bx}(t)}{\sqrt{\sigma^2 + 1}}, t\right) \diff \sigma(t).\label{eq:ddim_ode}
\vspace{-2mm}
\end{equation}
Then the inversion process(from $\bx_t$ to $\bx_{t+1}$) can be the reversion of the reconstruction process:
\vspace{-2mm}
\begin{equation}
    \frac{\bx_{t+1}}{\sqrt{\alpha_{t+1}}}  = \frac{\bx_t}{\sqrt{\alpha_t}}  + \left(\sqrt{\frac{1 - \alpha_{t+1}}{\alpha_{t+1}}} - \sqrt{\frac{1 - \alpha_{t}}{\alpha_t}}\right) \bepsilon_\theta(\bx_t, t). \label{eq:ddim_forward}
\vspace{-2mm}
\end{equation}
This process is to obtain the corresponding noisy sample $\bx_T$ for an input image $\bx_0$. However, it is very slow to invert or sample step by step. To speed up the diffusion model sampling, DDIM~\cite{DDIM} permits us  to sample a subset of $S$ steps $\tau_1, \dots, \tau_S$, so that the neighboring $\bx_t$ and $\bx_{t+1}$ become $\bx_{\tau_t}$ and $\bx_{\tau_{t+1}}$, respectively, in Eqn.~\eqref{eq:ddim_forward} and Eqn.~\eqref{eq:ddim_reverse}.

\begin{figure}[t] 
    \centering 
    \includegraphics[width=\linewidth]{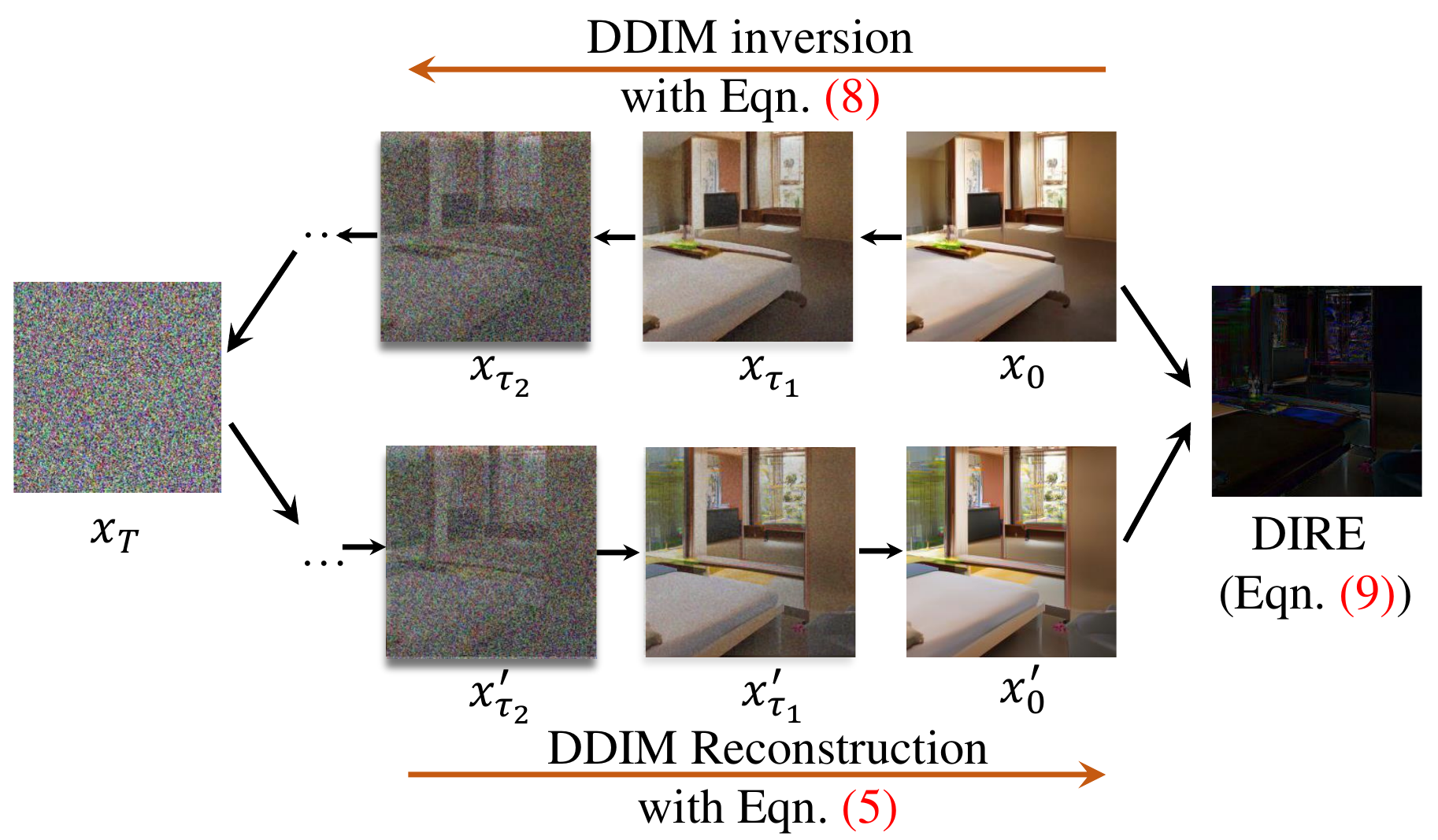}
    \vspace{-2em}
    \caption{\textbf{Illustration of the process of computing DIRE given an input image $\bx_0$}. The input image $\bx_0$ is first gradually inverted into a noise image $\bx_T$ by DDIM inversion~\cite{DDIM}, and then is denoised step by step until getting a reconstruction $\bx'_0$. DIRE is simply defined as the residual image got from $\bx_0$ and $\bx'_0$. }
    \label{fig:DIRE}
    \vspace{-1em}
\end{figure}

\subsection{DIRE}

Due to the intrinsic differences between diffusion models and previous generative models~(\ie, GANs, Flow-based models, VAEs), existing generated image detectors experience dramatic performance drops when facing images generated by diffusion models. To avoid the abuse of diffusion models, it is urgent to develop a detector for diffusion-generated image detection.
A straightforward approach would be to train a binary classifier using a dataset of both real and diffusion-generated images. However, it is difficult for such a method to guarantee generalization to diffusion models that have not been previously encountered.

Our research takes note of the fact that images generated by diffusion models are essentially sampled from the distribution of the diffusion generation space~($p_g(\bx)$), while real images are sampled from another distribution~($p_r(\bx)$) although it may be near to $p_g(\bx)$ but not exactly the same. Our core motivation is that samples from the diffusion generation space $p_g(\bx)$ are more likely to be reconstructed by a pre-trained diffusion model while real images cannot. 

So the key idea of our work is to make use of the diffusion model to detect diffusion-generated images. We find that images generated by diffusion models are more likely to be reconstructed by a pre-trained diffusion model. On the other hand, due to the complex characteristics of real images, real images can not be well constructed. As shown in Figure~\ref{fig:teaser}, the reconstruction error of real and diffusion-generated images shows dramatically different properties.

\begin{table}[t]
    \centering
    \resizebox{1.0\linewidth}{!}{
    \begin{tabular}{lclc}
 \hline
\multirow{2}{*}{Image Source} & Denoising & \multirow{2}{*}{Generator} & \# of images \\ 
& Condition & & (real/generated) \\ 
\hline
& \multirow{4}{*}{Unconditional} &ADM~\cite{ADM} & 42k/42k \\
& & DDPM~\cite{DDPM} & 42k/42k \\
& & iDDPM~\cite{iDDPM} & 42k/42k \\
LSUN& & PNDM~\cite{PNDM} & 42k/42k \\ 
\cline{2-4}
-Bedroom~\cite{yu2015lsun} & \multirow{4}{*}{Text2Image} &LDM~\cite{LDM} & 1k/1k \\
& & SD-v1~\cite{LDM} & 1k/1k \\
& & SD-v2~\cite{LDM} & 1k/1k \\
& & VQ-Diffusion~\cite{vq-diffusion} & 1k/1k \\ \hline
\multirow{2}{*}{ImageNet~\cite{deng2009imagenet}} & Conditional & ADM~\cite{ADM} & 50k/50k \\  \cline{2-4}
& Text2Image & SD-v1~\cite{LDM} & 10k/10k \\
\hline
\end{tabular}
}
\vspace{-.5em}
\caption{\textbf{Composition of the DiffusionForensics dataset.} It includes real images from LSUN-Bedroom~\cite{yu2015lsun} and ImageNet~\cite{deng2009imagenet}, and generated images from corresponding pre-trained diffusion models. According to the class of diffusion models, the containing images are divided into three classes: unconditional, conditional, and text2image.}
\vspace{-1em}
\label{tab:composition_dataset}
\end{table}

Given an input image $\bx_0$, we wish to judge whether it is synthesized by diffusion models. Take a pre-trained diffusion model $\bepsilon_\theta(\bx_t, t)$. As shown in Figure~\ref{fig:DIRE}, we apply the DDIM~\cite{DDIM} inversion process to gradually add Gaussian noise into $\bx_0$ via Eqn.~\eqref{eq:ddim_forward}.  After $S$ steps, $\bx_0$ becomes a point $\bx_T$ in the isotropic Gaussian noise distribution. The inversion process is to find the corresponding point in noisy space, then the DDIM~\cite{DDIM} generation process~(Eqn.~\eqref{eq:ddim_reverse}) is employed to reconstruct the input image and produces a recovered version $\bx'_0$. 
The differences between $\bx_0$ and $\bx'_0$ help to distinguish real or generated. Then the DIRE is defined as:
\vspace{-2mm}
\begin{equation}
    \text{DIRE}(\bx_0) = |\bx_0-\bR(\bI(\bx_0))|,
\vspace{-2mm}
\end{equation}
where $|\cdot|$ denotes computing the absolute value, $\bI(\cdot)$ is a series of the inversion process with Eqn.~\eqref{eq:ddim_forward} and $\bR(\cdot)$ is a series of the reconstruction process with Eqn.~\eqref{eq:ddim_reverse}. 

Then for real images and diffusion-generated images, we can get their DIRE representations, we train a binary classifier to distinguish their DIREs by a simple binary cross-entropy loss, which is formulated as follows,
\vspace{-3mm}
\begin{equation}
    L(\by, \by') = -\sum_{i=1}^N (\by_{i}\mathrm{log}(\by'_{i})+(1-\by_{i})\mathrm{log}(1-\by'_{i})),\label{eq:binary_loss}
\vspace{-3mm}
\end{equation}
where $N$ is mini-batch size, $\by$ is the ground-truth label, and $\by'$ is the corresponding prediction by the detector. In the inference stage, we first apply a diffusion model to reconstruct the image and get the DIRE. Subsequently, we input the DIRE into the binary classifier, which will then classify the source image as either real or generated.

\begin{table*}[t]
\small
\centering
\setlength\tabcolsep{2pt}
\resizebox{1.\linewidth}{!}{
\begin{tabular}{lccc cccccccc c}
\hline
\multirow{2}{*}{Method} & Training & Generation & Reconstruction & \multicolumn{8}{c}{Testing diffusion generators} & Total \\ \cline{5-12}
& dataset & model & model & ADM & DDPM & iDDPM& PNDM & SD-v1 & SD-v2 & LDM & VQDiffusion & Avg. \\ \hline
CNNDet~\cite{CNNDetection} & LSUN & ProGAN & -- & 50.1/63.4 & 56.7/74.6 & 50.1/77.6 & 50.3/82.9 & 50.2/70.9 & 50.8/80.4 &50.1/60.2 & 50.1/70.6 & 51.1/72.6 \\
GANDet~\cite{mandelli2022detecting} & LSUN & ProGAN  & -- & 54.2/43.6 & 52.2/47.3 & 45.7/57.3 & 42.1/77.6 & 68.1/78.5 & 61.5/52.7 &79.2/57.1 & 64.8/52.3 & 58.5/58.3\\ 
Patchfor~\cite{patchforensics} & FF++ & Multiple & -- & 50.4/74.8 & 56.8/67.4 & 50.3/69.5 & 55.1/78.5 & 49.9/84.7 & 50.0/52.8 &54.0/92.0 & 92.8/99.7 & 57.4/77.4 \\
SBI~\cite{SBI} & FF++ & Multiple & -- & 53.6/57.7 & 55.8/47.4 & 54.0/58.2 & 46.7/44.8 & 65.6/75.9 & 55.0/59.8 &81.0/88.3& 59.6/66.6 & 58.9/62.3\\
CNNDet*~\cite{CNNDetection} & LSUN-B. & ADM & -- & \textbf{100}/\textbf{100} & 87.3/99.6 & \textbf{100}/\textbf{100} & 77.8/99.1 & 77.4/85.8 & 83.4/98.2 & 96.0/99.9 & 70.3/96.1 & 86.5/97.3\\
Patchfor*~\cite{patchforensics} & LSUN-B. & ADM & -- & \textbf{100}/\textbf{100} & 72.9/\textbf{100} & \textbf{100}/\textbf{100} & 96.6/\textbf{100} & 63.2/71.3 & 97.2/\textbf{100} & 97.3/\textbf{100} & \textbf{100}/\textbf{100} & 90.9/96.4\\
F3Net*~\cite{qian2020thinking} & LSUN-B. & ADM & -- & 94.3/99.6 & 93.7/99.6 & 94.8/99.9 & 94.1/99.2 & 86.1/95.3 & 83.6/91.7 & 92.5/97.8 & 93.4/98.7 & 91.6/97.7 \\ \hline
\multirow{4}{*}{DIRE~(ours)} & LSUN-B. & ADM & ADM & \textbf{100}/\textbf{100} & \textbf{100}/\textbf{100} & \textbf{100}/\textbf{100} & 99.7/\textbf{100} & \textbf{99.7}/\textbf{100} & \textbf{100}/\textbf{100} & \textbf{100}/\textbf{100} & \textbf{100}/\textbf{100} & \textbf{99.9}/\textbf{100}\\ 
 & LSUN-B. & PNDM & ADM & \textbf{100}/\textbf{100} & \textbf{100}/\textbf{100} & \textbf{100}/\textbf{100} & \textbf{100}/\textbf{100} & 89.4/99.9 & \textbf{100}/\textbf{100} & \textbf{100}/\textbf{100} & \textbf{100}/\textbf{100} & 98.7/\textbf{100}\\
 & LSUN-B. & iDDPM & ADM & 99.6/100 & \textbf{100}/\textbf{100} & \textbf{100}/\textbf{100} & 89.7/99.8  & 99.7/\textbf{100} & \textbf{100}/\textbf{100}  & 99.9/\textbf{100} & 99.9/\textbf{100} & 98.6/\textbf{100} \\
 & LSUN-B. & StyleGAN & ADM & 98.8/\textbf{100} & 99.8/\textbf{100} & 99.9/\textbf{100} & 89.6/\textbf{100} & 95.2/\textbf{100} & \textbf{100}/\textbf{100} & \textbf{100}/\textbf{100} & \textbf{100}/\textbf{100}  & 97.9/\textbf{100}\\
\hline
\end{tabular}
}
\vspace{-.5em}
\caption{\textbf{Comprehensive comparisons of our DIRE and other generated image detectors.} The previous detectors including CNNDet~\cite{CNNDetection}, GANDet~\cite{mandelli2022detecting}, Patchfor~\cite{patchforensics}, and SBI~\cite{SBI} are tested on our DiffusionForensics benchmark using their provided models.  *~denotes our reproduction training on the ADM subset of DiffusionForensics with the official codes. 
All the used diffusion-generation models~\cite{ADM,PNDM,iDDPM} for preparing training data are unconditional models pre-trained on LSUN-Bedroom~(LSUN-B.)~\cite{yu2015lsun}. Generated images from StyleGAN~\cite{stylegan} on LSUN-Bedroom are downloaded from the official repository.
All the testing images produced by text-image generators~(SD-v1~\cite{LDM}, SD-v2~\cite{LDM}, LDM~\cite{LDM}, VQDiffusion~\cite{vq-diffusion}) are prompted by ``A photo of bedroom''. We report ACC~(\%) and AP~(\%)~(ACC/AP in the Table).}
\label{tab:mainresults}
\end{table*}

\subsection{DiffusionForensics: A Dataset for Evaluating Diffusion-Generated Image Detectors}
To better evaluate the performance of diffusion-generated detectors, we collect a dataset, DiffusionForensics,  which is comprised of ten subsets for comprehensive experiments.
Its composition is shown in Table~\ref{tab:composition_dataset}. The images can be roughly divided into two classes by their source: LSUN-Bedroom~\cite{yu2015lsun} and ImageNet~\cite{deng2009imagenet}. 

\noindent\textbf{LSUN-Bedroom.} We collect images generated by eight diffusion models trained on LSUN-Bedroom, in which four subset images~(ADM~\cite{ADM}, DDPM~\cite{DDPM}, iDDPM~\cite{iDDPM}, PNDM~\cite{PNDM}) are generated by unconditional diffusion models and the other four~(LDM~\cite{LDM}, SD-v1~\cite{LDM}, SD-v2~\cite{LDM}, VQ-Diffusion~\cite{vq-diffusion}) are generated by text2image diffusion models. The text prompt for all the text2image generation is ``A photo of bedroom''. 
For the images generated by unconditional diffusion models, the generated 42k images in ADM~\cite{ADM}, iDDPM~\cite{iDDPM}, and PNDM~\cite{PNDM} are split into 40k~(training), 1k~(validation), and 1k~(testing). The 1k images generated by DDPM~\cite{DDPM} are used for testing. Further for generalization evaluation to text2image generation, 1k images generated by using pre-trained LDM~\cite{LDM}, SD-v1~\cite{LDM}, SD-v2~\cite{LDM}, and VQ-Diffusion~\cite{vq-diffusion} models or provided APIs are generated, respectively.

\noindent\textbf{ImageNet.} We further collect images from ImageNet for evaluating detectors when facing more universal image generation and cross-dataset evaluation. To be specific, we collect images from a conditional diffusion model~(ADM~\cite{ADM}), and a text2image diffusion model~(SD-v1~\cite{LDM}) in which the text prompt for generation is ``A photo of \{class\}''(1k classes from ImageNet~\cite{deng2009imagenet}). Applying the pre-trained ADM model~\cite{ADM} with classifier, we generate 50k images in total~(50 images for each class in ImageNet), \ie, 40k for training, 5k for validation, and 5k for testing.
And images generated by the text2image model~\cite{LDM} are only for testing.

The split of real images for training/validation/testing is the same as the corresponding generated images. Besides, all the data in our dataset are triplet, \ie, source image, reconstructed image, and corresponding DIRE image. In general, the proposed DiffusionForensics dataset contains unconditional, conditional, and text2image generated images, which is fertile for evaluation from various aspects.

\section{Experiment}

In this section, we first introduce the experimental
setups and then provide extensive experimental results to
demonstrate the superiority of our approach.

\subsection{Experimental Setup}
\noindent\textbf{Data pre-processing and augmentation.}
All the experiments are conducted on our DiffusionForensics dataset. To calculate the DIRE for each image, we use the ADM~\cite{ADM} network pre-trained on LSUN-Bedroom as the reconstruction model, and the DDIM~\cite{DDIM} inversion and reconstruction process in which the number of steps $S=20$ by default. We employ ResNet-50~\cite{resnet} as our forensics classifier. The size of most images~(ADM~\cite{ADM}, DDPM~\cite{DDPM}, iDDPM~\cite{iDDPM}, PNDM~\cite{PNDM}, VQ-Diffusion~\cite{vq-diffusion}, LDM~\cite{LDM}) in the dataset is 256 $\times$ 256. For Stable Diffusion~\cite{LDM} v1 and v2, the generated images are resized into 256 $\times$ 256 with bicubic interpolation. During training, the images fed into the network are randomly cropped with the size of 224 $\times$ 224 and horizontally flipped with a probability of 0.5. 
During testing, the images are center-cropped with the size of 224 $\times$ 224.

\noindent\textbf{Evaluation metrics.} Following previous generated-image detection methods~\cite{CNNDetection,wang2019detecting,zhou2018learning}, we mainly report accuracy~(ACC) and average precision~(AP) in our experiments to evaluate the detectors. The threshold for computing accuracy is set to 0.5 following \cite{CNNDetection}. 

\noindent\textbf{Baselines.} 1) CNNDetection~\cite{CNNDetection} proposes a CNN-generated image detection model that can be trained on one CNN dataset and then generalized to other CNN-synthesized images. 2) GANDetection~\cite{mandelli2022detecting} applies an ensemble of EfficientNet-B4~\cite{tan2019efficientnet} to increase the detection performance. 3) SBI~\cite{SBI} trains a general synthetic-image detector on images generated by blending pseudo source and target images from single pristine images. 4) Patchforensics~\cite{patchforensics} employs a patch-wise classifier which is claimed to be better than simple classifiers for fake image detection. 5) F3Net~\cite{qian2020thinking} proposes that the frequency information of images is essential for fake image detection.

\subsection{Comparison to Existing Detectors} 
Diffusion models~\cite{DDPM,ADM} are claimed to exhibit better generation ability than previous generation models~(\eg, GAN~\cite{goodfellow2014generative}, VAE~\cite{kingma2013auto}). We notice that previous detectors achieve surprising performance on images generated by CNNs~\cite{stylegan,choi2018stargan,biggan}, but the generalization ability when facing recent diffusion-generated images has not been well explored. Here, we evaluate CNNDetection~\cite{CNNDetection}, GANDetection~\cite{mandelli2022detecting}, Patchforensics~\cite{patchforensics}, 
 and SBI~\cite{SBI} on the proposed DiffusionForensics dataset using the pre-trained weights downloaded from their official repositories.

The quantitative results can be found in Table~\ref{tab:mainresults}. We find that existing detectors have a significant performance drop in performance when dealing with diffusion-generated images, with ACC results lower than 60\%. We also include diffusion-generated images~(ADM~\cite{ADM}) as training data and re-train CNNDetection~\cite{CNNDetection}, Patchforensics~\cite{patchforensics}, and F3Net~\cite{qian2020thinking}, whose training codes are publicly available. The resulting models get a significant improvement on images generated by the same diffusion models as used in training, but still perform unsatisfactorily facing unseen diffusion models. In contrast, our method, DIRE, shines with excellent generalization performance. Concretely, DIRE with the generation model and the reconstruction model set to ADM achieves an average of 99.9\%  ACC and 100\% AP on detecting images generated by various diffusion models.

\begin{table}[t]
    \centering
    \setlength\tabcolsep{2pt}
    \resizebox{1.0\linewidth}{!}{
    \begin{tabular}{lcc@{\hskip 4mm}rc}
\hline
Training & Generation  & \multicolumn{3}{c}{Testing generators}  \\\cline{3-5}
dataset & model & ADM(IN) & SD-v1(IN) & StyleGAN(LSUN-B.) \\ \hline
LSUN-B. & ADM & \textbf{90.2}/\textbf{97.9} & \textbf{97.2}/\textbf{99.8}  & 99.9/\textbf{100}\\ 
LSUN-B. & iDDPM & \textbf{90.2}/\textbf{97.9} & 93.7/99.3 & 99.9/\textbf{100} \\ 
LSUN-B. & StyleGAN & 76.9/94.4 & 89.7/99.0 & \textbf{100}/\textbf{100}\\
\hline
\end{tabular}
}
\vspace{-.5em}
\caption{\textbf{Cross-dataset evaluation} on ImageNet~(IN)~\cite{deng2009imagenet} and LSUN-bedroom~(LSUN-B.)~\cite{yu2015lsun}. Each testing generator is pre-trained on the corresponding dataset. Images generated by Stable Diffusion-v1~(SD-v1) are prompted by `` A photo of \{class\}'' in which the classes are from~\cite{deng2009imagenet}. ACC~(\%) and AP~(\%) are reported~(ACC/AP in the Table).}
\vspace{-1em}
\label{tab:eval_imagenet}
\end{table}

\subsection{Generalization Capability Evaluation}
\noindent\textbf{Effect of choice of generation and reconstruction models.} 
We evaluate the impact of different choices of the generation and reconstruction models on the generalization capability. We employ the ADM~\cite{ADM} model as the reconstruction model and apply different models for generating images. After generating, the ADM model converts these images to their DIREs for training a binary classifier. In this evaluation, we select three different generation models: PNDM~\cite{PNDM} and iDDPM~\cite{iDDPM}~(diffusion models) and StyleGAN~\cite{stylegan}~(GAN model). The results are reported in Table~\ref{tab:mainresults}. Despite the inconsistent use of generation and reconstruction models when training, DIRE still keeps a strong generalization capability. Specifically, when pairing iDDPM~\cite{iDDPM} as the generation model and ADM~\cite{ADM} as the reconstruction model, DIRE achieves 98.6\% ACC and 100\% AP on average, highlighting its adaptation with images generated by different diffusion models. It's worth noting that when the generation model is StyleGAN, DIRE still exhibits excellent performance. This might be attributed to DIRE's capability of incorporating the generation properties of other generation models besides diffusion models.

\begin{figure*}[t]
    \includegraphics[width=\linewidth]{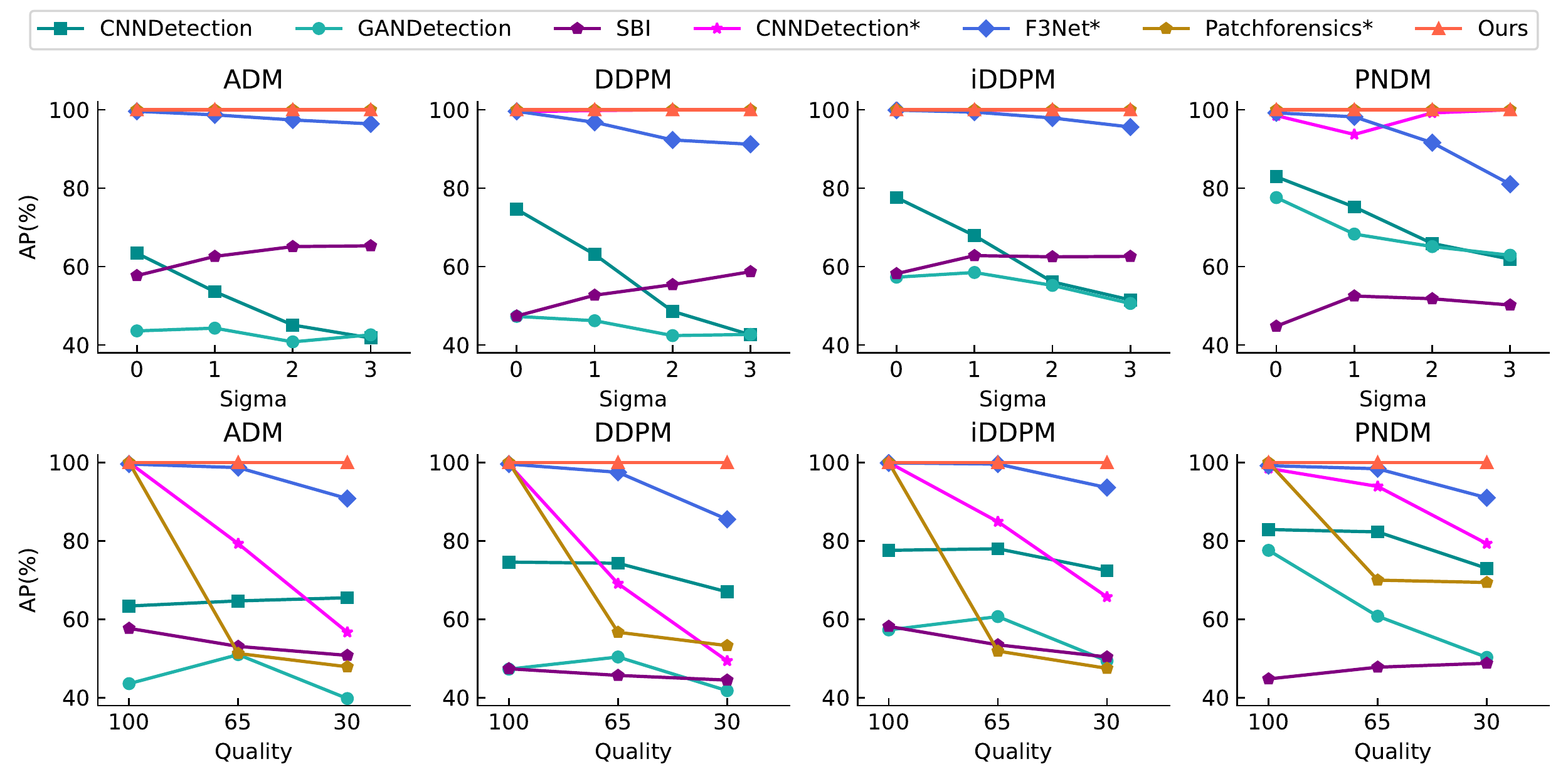}
\vspace{-2.5em}
\caption{\textbf{Robustness to unseen perturbations.} The top rows show the robustness to Gaussian blur, and the bottom rows show the robustness to JPEG compression. 
* denotes our reproduction training on the ADM subset of DiffusionForensics with AP~(\%) reported for robustness comparison.}
\vspace{-1em}
\label{fig:rubostness}
\end{figure*}

\noindent\textbf{Cross-dataset evaluation.} We further design a more challenging scenario, \ie, training the detector with images generated by models pre-trained on LSUN-Bedroom~\cite{yu2015lsun} and then testing it on images produced by models pre-trained on ImageNet~\cite{deng2009imagenet}. We choose three different generators for generating training images: ADM~\cite{ADM}, iDDPM~\cite{iDDPM}, and StyleGAN~\cite{stylegan}. The evaluation results on ADM~(IN) are shown in Table~\ref{tab:eval_imagenet}. The comparison indicates that DIRE maintains a satisfactory generalization capability even though facing unseen datasets, \ie, ACC/AP: 90.2\%/97.9\% when training on images generated by ADM and iDDPM. And for StyleGAN~\cite{stylegan}, DIRE still achieves 94.4\% in AP, although there is a huge domain gap that both the dataset and generation models are different in training and testing.  This evaluation further validates that the proposed DIRE is a general image representation for this task.

\noindent\textbf{Unseen text-to-image generation evaluation.} Furthermore, we seek to verify whether DIRE can detect images generated by unseen text-to-image models. We adopt Stable Diffusion v1~(SD-v1) as the generation model and generate images based on the class label of ImageNet~\cite{deng2009imagenet} for evaluating detectors. The results are shown in Table~\ref{tab:eval_imagenet}. Our detector trained with images generated by ADM pre-trained on LSUN-Bedroom achieves a 97.2\% ACC and 99.8\% AP, demonstrating the strong generalization capability of DIRE to text-to-image generation models.   

\noindent\textbf{Unseen GAN generation evaluation.}  
Besides generalization between diffusion models, we further evaluate the performance of DIRE for images generated by GANs. We test the performance of three generation models: ADM, iDDPM, and StyleGAN pre-trained on LSUN-Bedroom. The results are reported in Table~\ref{tab:eval_imagenet}. Although the classifier never encounters any GAN-generated image during training, it achieves surprising performance when detecting GAN-generated images, \ie, 99\% in ACC and 100\% in AP. This indicates that DIRE is not only an effective image representation for diffusion-generated image detection but also may beneficial to detect GAN-generated images.

\begin{table}[t]
    \centering
    \resizebox{1.0\linewidth}{!}{
    \begin{tabular}{cccccc}
\hline
$S$ & ADM  & DDPM   & iDDPM   & PNDM   & SD-v1 \\ \hline
5 & \textbf{100}/\textbf{100} & \textbf{100}/\textbf{100} & \textbf{100}/\textbf{100} & 97.5/\textbf{100} & 87.5/99.8 \\
10 & \textbf{100}/\textbf{100} & \textbf{100}/\textbf{100} & \textbf{100}/\textbf{100} & 99.4/\textbf{100} & 98.2/\textbf{100} \\
20 & \textbf{100}/\textbf{100} & \textbf{100}/\textbf{100} & \textbf{100}/\textbf{100} & 99.7/\textbf{100} & 99.7/\textbf{100} \\ 
\textbf{50} & \textbf{100}/\textbf{100} & \textbf{100}/\textbf{100} & \textbf{100}/\textbf{100} & \textbf{100}/\textbf{100} & \textbf{99.9}/\textbf{100} \\ 
\hline
\end{tabular}
}
\vspace{-.5em}
\caption{\textbf{Influence of different inversion steps.} All the models in this experiment are trained on the ADM subset. ACC~(\%) and AP~(\%) are reported~(ACC/AP in the Table).}
\vspace{-1em}
\label{tab:different_steps}
\end{table}

\subsection{Robustness to Unseen Perturbations}
Besides the generalization to unseen generation models, the robustness to unseen perturbations is also a common concern since in real-world applications images are usually perturbed by various degradations. Here, we evaluate the robustness of detectors in two-class degradations, \ie, Gaussian blur and JPEG compression, following~\cite{CNNDetection}. The perturbations are added under three levels for Gaussian blur~($\sigma=1, 2, 3$) and two levels for JPEG compression~($quality=65, 30$). We explore the robustness of our baselines CNNDetection~\cite{CNNDetection}, GANDetection~\cite{mandelli2022detecting}, SBI~\cite{SBI}, F3Net~\cite{qian2020thinking}, Patchforensics~\cite{patchforensics}, and our DIRE. The results are shown in Figure~\ref{fig:rubostness}. We observe that at each level of blur and JPEG compression, our DIRE gets a perfect performance without performance drop.
It is worth noting that our reproduction of CNNDetection*~\cite{CNNDetection} and Patchforensics*~\cite{patchforensics} trained on the ADM subset of DiffusionForensics also get satisfactory performance while they experience a dramatic performance drop facing JPEG compression, which further reveals training on RGB images may be not robust.

\begin{table}[t]
    \centering
    \setlength\tabcolsep{2pt}
    \resizebox{1.0\linewidth}{!}{
    \begin{tabular}{cccccc}
\hline
Input & ADM  & DDPM  & iDDPM   & PNDM   & SD-v1  \\ \hline
REC & \textbf{100}/\textbf{100} & 57.1/57.7 & 49.7/92.6 & 87.1/98.7 & 46.9/57.0 \\
RGB & \textbf{100}/\textbf{100} & 87.3/99.6 & \textbf{100}/\textbf{100}& 77.8/99.1&
77.4/85.8 \\
RGB\&DIRE & \textbf{100}/\textbf{100} & 99.8/\textbf{100}& 99.9/\textbf{100}& 99.2/\textbf{100}&
62.4/92.4 \\
\textbf{DIRE} & \textbf{100}/\textbf{100} & \textbf{100}/\textbf{100} & \textbf{100}/\textbf{100} & \textbf{99.7}/\textbf{100} & \textbf{99.7}/\textbf{100} \\ \hline
\end{tabular}
}
\vspace{-.5em}
\caption{\textbf{Influence of different input information.} All the models in this experiment are trained on the ADM subset. ACC~(\%) and AP~(\%) are reported~(ACC/AP in the Table).}
\vspace{-.5em}
\label{tab:different_input}
\end{table}

\begin{table}[t]
    \centering
    \setlength\tabcolsep{2pt}
    \resizebox{1.0\linewidth}{!}{
    \begin{tabular}{cccccc}
\hline
  & ADM  & DDPM   & iDDPM   & PNDM   & SD-v1  \\ \hline
w/o ABS & \textbf{100}/\textbf{100} & 99.4/\textbf{100} & \textbf{100}/\textbf{100} &98.2/\textbf{100} & 87.0/93.0 \\
\textbf{w/ ABS} & \textbf{100}/\textbf{100} & \textbf{100}/\textbf{100} & \textbf{100}/\textbf{100} & \textbf{99.7}/\textbf{100} & \textbf{99.7}/\textbf{100} \\
\hline
\end{tabular}
}
\vspace{-.5em}
\caption{\textbf{Effect of computing the absolute value~(ABS) when obtaining DIRE.} All the models in this experiment are trained on the ADM subset. ACC~(\%) and AP~(\%) are reported~(ACC/AP in the Table).}
\vspace{-1em}
\label{tab:effect_abs}
\end{table}

\begin{figure*}[t] 
    \centering 
    \includegraphics[width=\linewidth]{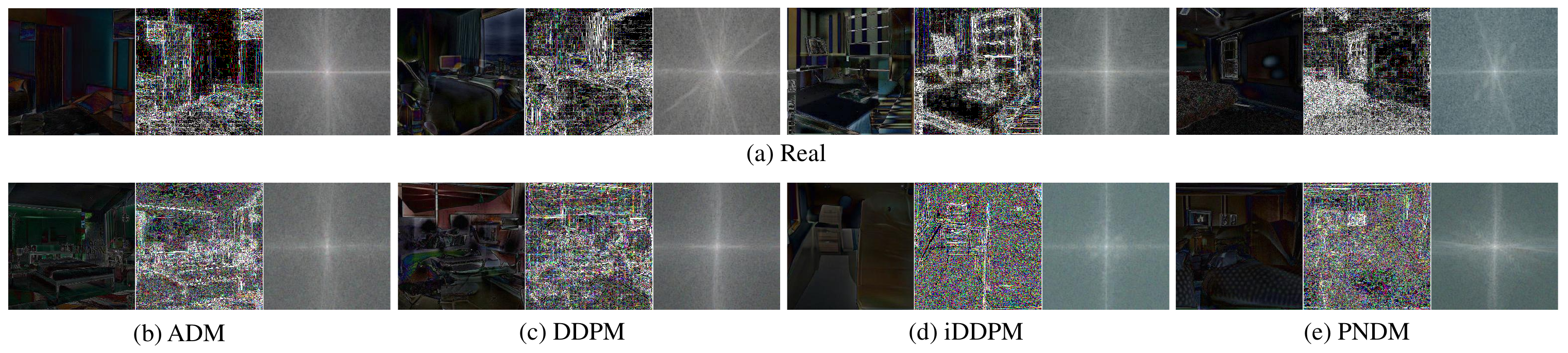}
    \vspace{-2em}
    \caption{\textbf{Noise pattern and frequency analysis of DIRE of real and generated images.} Noise pattern is regular to portray the shape of objects in DIRE of real images, while it is messy in DIRE of diffusion-generated images. For frequency analysis, the frequency bands in DIRE of real images are more abundant than that of diffusion-generated images, \ie, the white regions in the frequency domain are larger.}
    \vspace{-1em}
    \label{fig:noise_fft}
\end{figure*}

\subsection{More Analysis of the Proposed DIRE}
\noindent \textbf{How do the inversion steps in DDIM affect the detection performance?} Recent diffusion models~\cite{DDIM,ADM} find that more steps contribute to more high-quality images and DDIM~\cite{DDIM} sampling can improve the generation performance compared to original DDPM~\cite{DDPM} sampling. Here, we explore the influence of different inversion steps in diffusion-generated image detection. Note that the steps in reconstruction are the same as in the inversion by default. The results are reported in Table~\ref{tab:different_steps}. We observe that more steps in DDIM benefit the detection performance of DIRE. Considering the computational cost, we choose 20 steps by default.  

\noindent \textbf{Is DIRE really better than the original RGB for detecting diffusion-generated images?} We conduct an experiment on various forms of input for detection, including RGB images, reconstructed images (REC), DIRE, and the combination of RGB and DIRE (RGB\&DIRE). The results displayed in Table~\ref{tab:different_input} reveal that REC performed much worse than RGB, suggesting that reconstructed images are not suitable as input information for detection. One possible explanation is the loss of essential information during reconstruction by a pre-trained diffusion model. The comparison between RGB and DIRE also demonstrates that DIRE serves as a stronger image representation, contributing to a more generalizable detector than simply training on RGB images. Furthermore, we find that combining RGB with DIRE together hurts the generalization compared to pure DIRE. Therefore, we use DIRE as the default input for detection by default.

\noindent \textbf{Effect of different calculation of DIRE.} After computing the residual result of the reconstructed image and source image, whether to compute the absolute value should be considered. As reported in Table~\ref{tab:effect_abs}, we find that the absolute operation is critical for achieving a strong diffusion-generated image detector, particularly on SD-v1~\cite{LDM} where it improves ACC/AP from 87.0\%/93.0\% $\rightarrow$ 99.7\%/100\%. So by default, the absolute operation is applied in all our models.

\noindent \textbf{Qualitative Analysis of DIRE.}
The above quantitative experiments have indicated the effectiveness of the proposed DIRE. As analyzed before, the key motivation behind DIRE is that generated images can be approximately reconstructed by a pre-trained diffusion model while real images cannot. DIRE makes use of the residual characteristic of an input image and its reconstruction for discrimination. To gain a better understanding of its intrinsic properties, we conduct a further qualitative analysis of DIRE, utilizing noise pattern and frequency analysis for visualization.

When images are acquired, various factors from hardware facilities, such as lens and sensors, and software algorithms, such as compression and demosaic, can impact image quality at the low level. One typical low-level analysis of images is noise pattern analysis\footnote{https://29a.ch/photo-forensics/\#noise-analysis}, which is usually regular and corresponds to the shape of objects in real scenarios. In addition to low-level analysis, frequency analysis can provide frequency information about images. To compute the frequency information of DIRE, we used FFT algorithms.

We visualize the results of the aforementioned two analysis tools in Figure~\ref{fig:noise_fft}. The visual comparison of noise patterns highlights significant differences of the DIRE of real and diffusion-generated images from the low-level perspective, with real images tending to be regular and corresponding to the shape of objects while diffusion-generated images tend to be messy. By comparing the FFT spectrum of DIRE from real and diffusion-generated images, we observe that the FFT spectrum of real images is usually more abundant than that of diffusion-generated images, which confirms that real images are more difficult to be reconstructed by a pre-trained diffusion model.

\section{Conclusion}
In this paper, we focus on building a generalizable detector for discriminating diffusion-generated images. We find that previous generated-image detectors show limited performance when detecting images generated by diffusion models. To address the issue, we present an image representation called DIRE based on reconstruction errors of images inverted and reconstructed by DDIM. Furthermore, we create a new dataset, DiffusionForensics, which includes images generated by unconditional, conditional, and text-to-image diffusion models to facilitate the evaluation of diffusion-generated images. Extensive experiments indicate that the proposed image representation DIRE contributes to a strong diffusion-generated image detector, which is very effective for this task.
We hope that our work can serve as a solid baseline for diffusion-generated image detection.

{\small
\bibliographystyle{ieee_fullname}
\bibliography{egbib}
}

\clearpage
\newpage
\appendix
\section{More Details of DiffusionForensics}

In this section, we give more details about the proposed DiffusionForensics dataset. 
The real images in the LSUN-Bedroom and ImageNet subsets are from the source LSUN-Bedroom~\cite{yu2015lsun} and ImageNet~\cite{deng2009imagenet} datasets, respectively.
To generate DIREs of all the real and generated images in our DiffusionForensics, we use the unconditional ADM~\cite{ADM} model pre-trained on LSUN-Bedroom  with DDIM~\cite{DDIM} scheduler applied for 20 steps in total as the reconstruction model. As for the ImageNet subset, we also provide the DIREs computed by using the unconditional ADM~\cite{ADM} model pre-trained on ImageNet as the reconstruction model with DDIM~\cite{DDIM} scheduler applied for 20 steps in total.

\noindent\textbf{LSUN-Bedroom-ADM.} We download the pre-trained LSUN-Bedroom model of ADM~\cite{ADM} from the official repository\footnote{https://github.com/openai/guided-diffusion}. And then we sample 42k images for training~(40k), validation~(1k), and testing~(1k) with DDIM~\cite{DDIM} scheduler applied for better sampling with 50 steps.

\noindent\textbf{LSUN-Bedroom-DDPM.} We download the provided 1k images generated by the model pre-trained on LSUN-Bedroom from the official repository\footnote{https://github.com/hojonathanho/diffusion}. 

\noindent\textbf{LSUN-Bedroom-iDDPM.} We sample 42k images using the official codes and the pre-trained LSUN-Bedroom model~(lr$=$2e-5)\footnote{https://github.com/openai/improved-diffusion} with DDIM~\cite{DDIM} scheduler applied for better sampling with 50 steps. 

\noindent\textbf{LSUN-Bedroom-PNDM.} We sample 42k images using the official codes and the pre-trained LSUN-Bedroom model\footnote{https://github.com/luping-liu/PNDM} with their PNDM~\cite{PNDM} scheduler applied for better sampling with 50 steps. 

\noindent\textbf{LSUN-Bedroom-LDM.} The pipeline code for sampling is downloaded from diffusers~\cite{von-platen-etal-2022-diffusers}\footnote{https://github.com/huggingface/diffusers}. The version of the Latent Diffusion model~(LDM)~\cite{LDM} we used is ``CompVis/ldm-text2im-large-256''. We give a prompt ``A photo of bedroom'' for generating 1k bedroom images. 

\noindent\textbf{LSUN-Bedroom-SD-v1.} The pipeline code for sampling is downloaded from diffusers~\cite{von-platen-etal-2022-diffusers}. The version of Stable Diffusion~(SD) v1~\cite{LDM} we used is ``runwayml/stable-diffusion-v1-5''. We give a prompt ``A photo of bedroom'' for generating 1k bedroom images. 

\noindent\textbf{LSUN-Bedroom-SD-v2.} The pipeline code for sampling is downloaded from diffusers~\cite{von-platen-etal-2022-diffusers}. The version of Stable Diffusion~(SD) v2~\cite{LDM} we used is ``stabilityai/stable-diffusion-2''. We give a prompt ``A photo of bedroom'' for generating 1k bedroom images. 

\noindent\textbf{LSUN-Bedroom-VQDiffusion.} We sample 1k images using the official codes and the pre-trained ITHQ model\footnote{https://github.com/microsoft/VQ-Diffusion} with the prompt ``A photo of bedroom''.

\noindent\textbf{ImageNet-ADM.} We sample 50k images for 1k classes from ImageNet~\cite{deng2009imagenet} using the pre-trained conditional ADM model and the provided classifier on ImageNet from the official repository with DDIM~\cite{DDIM} scheduler applied for better sampling with 50 steps. The images are divided in the ratio of 8$:$1$:$1 for training$:$validation$:$testing.

\noindent\textbf{ImageNet-SD-v1.} We sample 10k images using the pre-trained Stable Diffusion v1.5 model with code provided by diffusers~\cite{von-platen-etal-2022-diffusers}. The prompt for generation is ``A photo of \{class\}'' in which the class is chosen from the 1k classes from ImageNet~\cite{deng2009imagenet}, resulting in ten generated images for each class.

\section{More Explanation of DIRE}

We have demonstrated the key motivation of our DIRE.
But one may wonder why DIREs of diffusion-generated images are not zero-value images. Here, we explain from the approximation to solving ordinary differential equations (ODEs) perspective.

The deterministic reverse process~(\textit{reconstruction}) in DDIM~\cite{DDIM} is as follows,
\begin{equation}
\begin{aligned}
    \bx_{t-1} = \sqrt{\alpha_{t-1}} \left(\frac{\bx_t - \sqrt{1 - \alpha_t} \bepsilon_\theta(\bx_t, t)}{\sqrt{\alpha_t}}\right) \\
    + \sqrt{1 - \alpha_{t-1}} \cdot \bepsilon_\theta(\bx_t, t).\label{eq:supp_ddim_reverse}
\end{aligned}
\end{equation}
  When the total steps $T$ is large enough~(\eg, $T=1000$), Eqn.~\eqref{eq:supp_ddim_reverse} can be seen as Euler integration for solving ordinary differential equations (ODEs):
\begin{equation}
    \frac{\bx_{t-\Delta t}}{\sqrt{\alpha_{t-\Delta t}}}  = \frac{\bx_t}{\sqrt{\alpha_t}}  + \left(\sqrt{\frac{1 - \alpha_{t-\Delta t}}{\alpha_{t-\Delta t}}} - \sqrt{\frac{1 - \alpha_{t}}{\alpha_t}}\right) \bepsilon_\theta(\bx_t, t). \label{eq:supp_ddim_euler}
\end{equation}

Suppose $\sigma=\sqrt{1-\alpha}/\sqrt{\alpha}$, $\bar\bx=\bx/\sqrt{\alpha}$, the corresponding ODE becomes:
\begin{equation}
    \diff \bar{\bx}(t) = \bepsilon_\theta\left(\frac{\bar{\bx}(t)}{\sqrt{\sigma^2 + 1}}, t\right) \diff \sigma(t).\label{eq:supp_ddim_ode}
\end{equation}
Then the forward process~(\textit{inversion})~(from $\bx_t$ to $\bx_{t+1}$) in DDIM can be the reversion of the reconstruction process:
\begin{equation}
    \frac{\bx_{t+1}}{\sqrt{\alpha_{t+1}}}  = \frac{\bx_t}{\sqrt{\alpha_t}}  + \left(\sqrt{\frac{1 - \alpha_{t+1}}{\alpha_{t+1}}} - \sqrt{\frac{1 - \alpha_{t}}{\alpha_t}}\right) \bepsilon_\theta(\bx_t, t). \label{eq:supp_ddim_forward}
\end{equation}

It is worth noting that during the approximation of Eqn.~\eqref{eq:supp_ddim_reverse} by Eqn.~\eqref{eq:supp_ddim_euler}, there is a deviation since $T$ is usually not infinitely large~(\eg, $T=1000$).  The deviation is more prominent for real images than diffusion-generated images due to the more complex characteristics of real images. The deviation caused by the approximation actually leads to our key idea of DIRE.

\section{More Visualization About DIRE}
\begin{figure*}[t] 
    \centering 
    \includegraphics[width=0.98\linewidth]{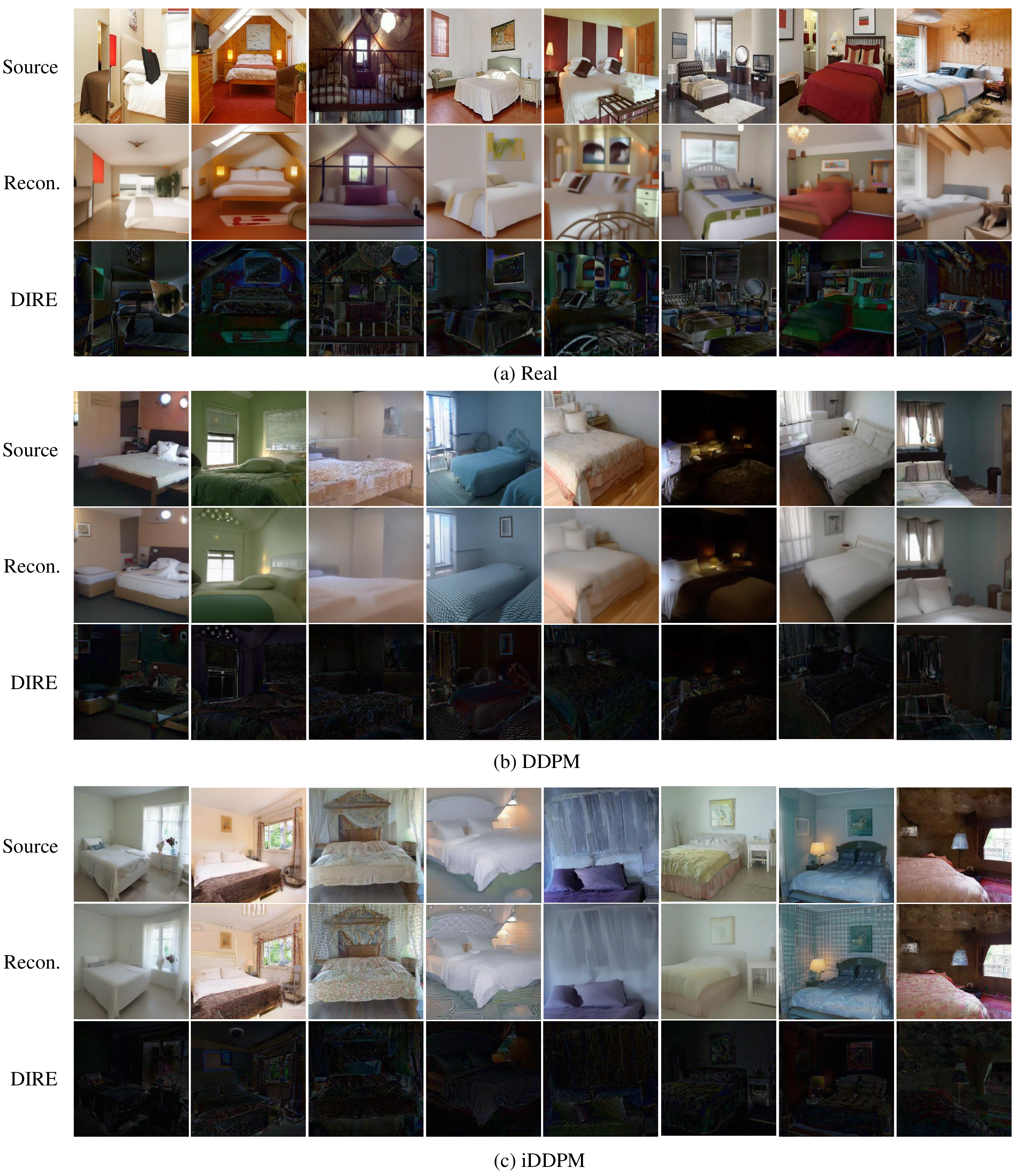}
    \vspace{-1em}
    \caption{The DIRE representation of real images and generated images from DDPM~\cite{DDPM} and iDDPM~\cite{iDDPM} pre-trained on LSUN-Bedroom~\cite{yu2015lsun}. For each source image, we visualize its corresponding reconstruction image and DIRE.}
    \label{fig:dire_lsun1}
    \vspace{-5.0mm}
\end{figure*}

\begin{figure*}[t] 
    \centering 
    \includegraphics[width=\linewidth]{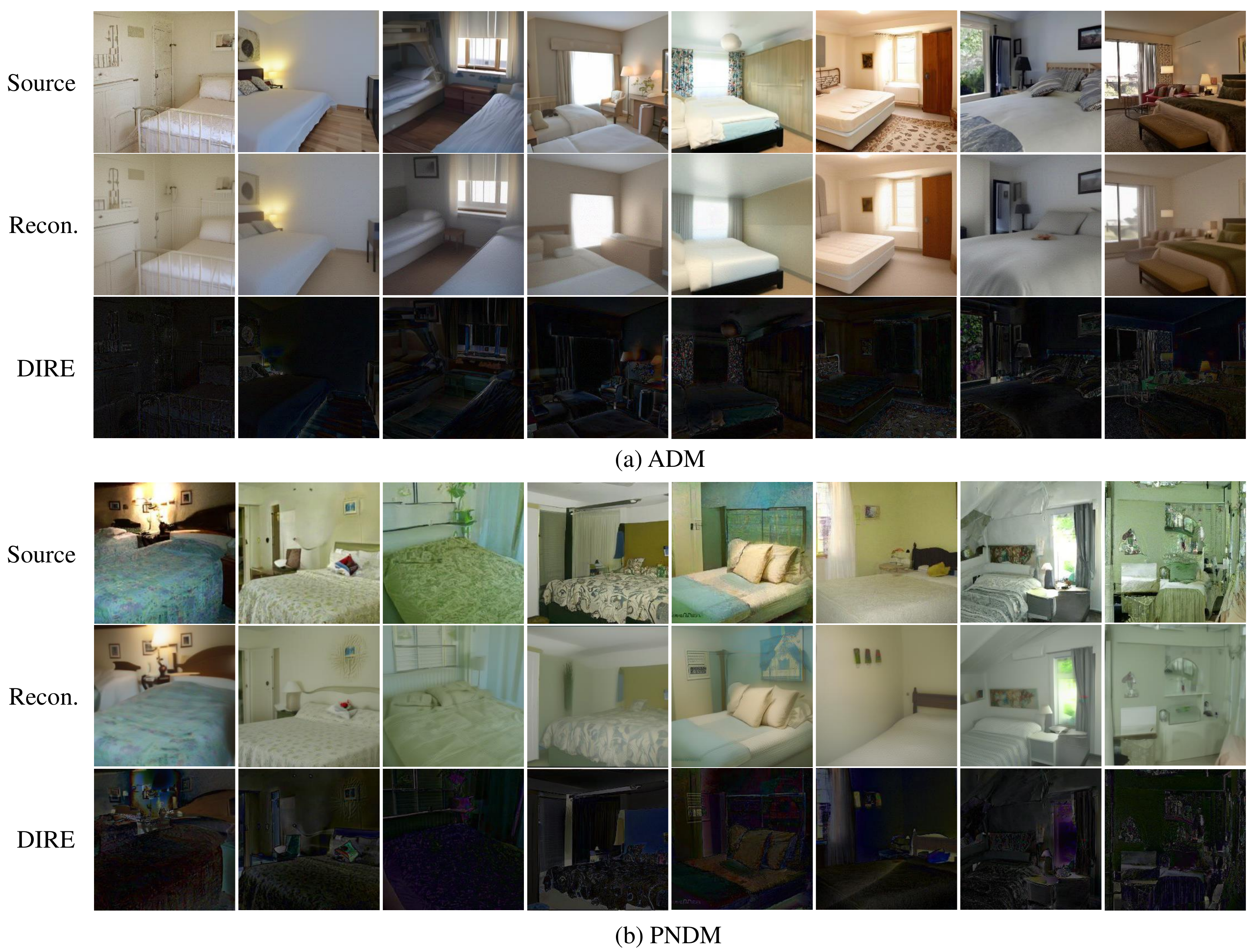}
    \vspace{-1em}
    \caption{The DIRE representation of generated images from ADM~\cite{ADM} and PNDM~\cite{PNDM} pre-trained on LSUN-Bedroom~\cite{yu2015lsun}. For each source image, we visualize its corresponding reconstruction image and DIRE.}
    \label{fig:dire_lsun2}
    \vspace{-5.0mm}
\end{figure*}

\begin{figure*}[t] 
    \centering 
    \includegraphics[width=\linewidth]{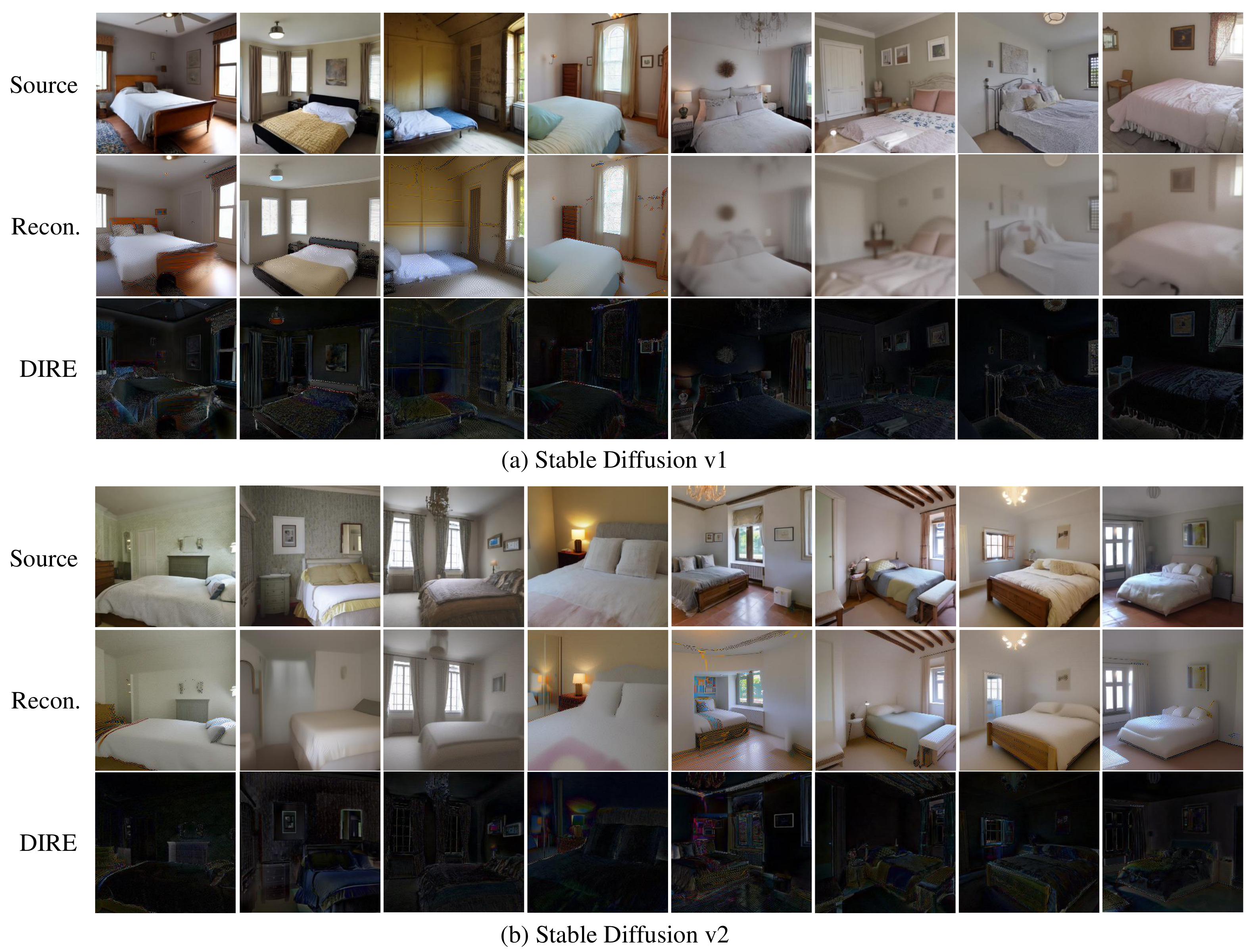}
    \vspace{-1em}
    \caption{The DIRE representation of generated images from Stable Diffusion v1 and v2~\cite{LDM} with the prompt ``A photo of bedroom''. For each source image, we visualize its corresponding reconstruction image and DIRE.}
    \label{fig:dire_lsun3}
    \vspace{-5.0mm}
\end{figure*}

\begin{figure*}[t] 
    \centering 
    \includegraphics[width=\linewidth]{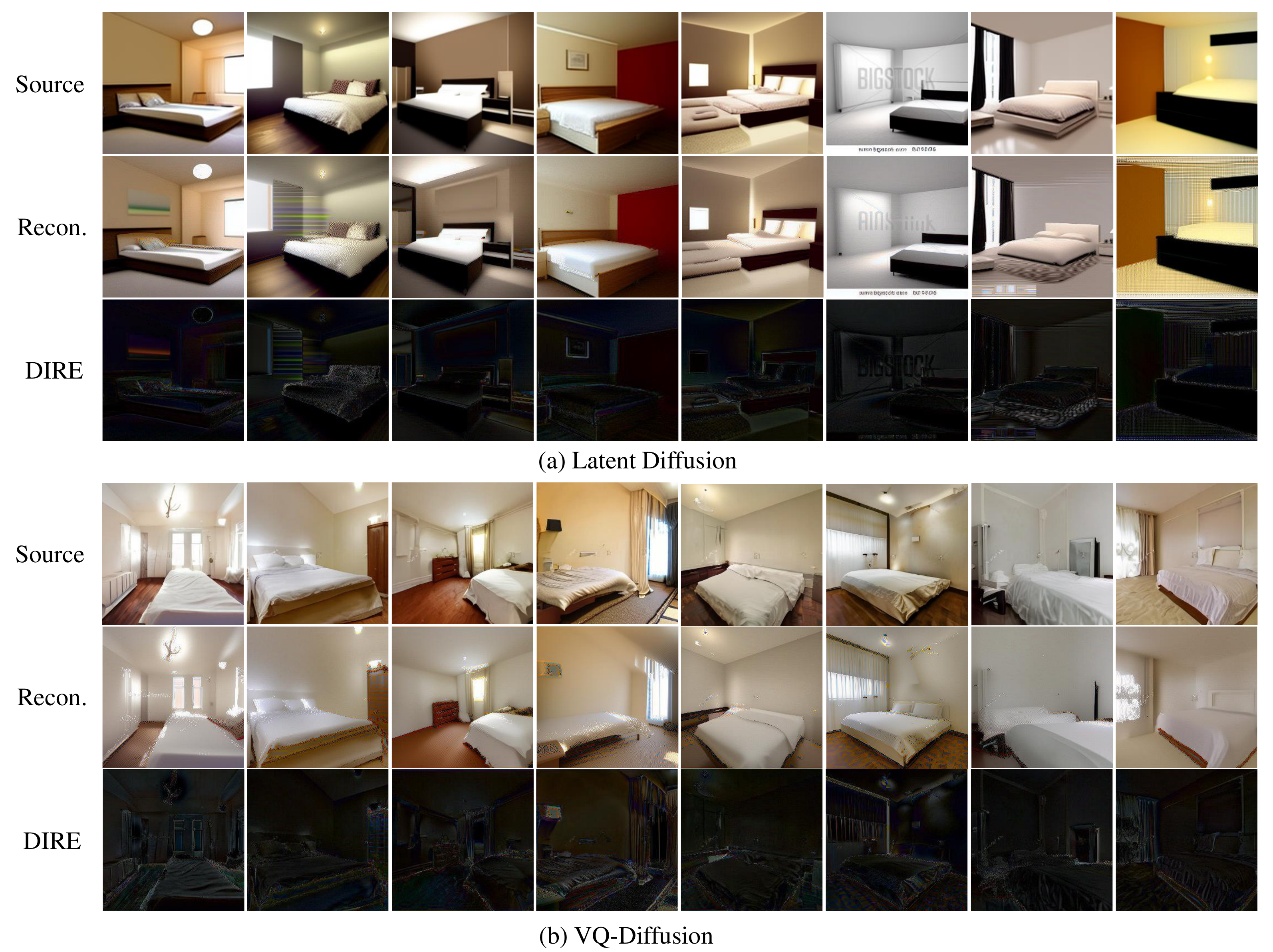}
    \vspace{-1em}
    \caption{The DIRE representation of generated images from Latent Diffusion~\cite{LDM} and VQ-Diffusion~\cite{vq-diffusion} with the prompt ``A photo of bedroom''. For each source image, we visualize its corresponding reconstruction image and DIRE.}
    \label{fig:dire_lsun4}
    \vspace{-5.0mm}
\end{figure*}

\begin{figure*}[t] 
    \centering 
    \includegraphics[width=\linewidth]{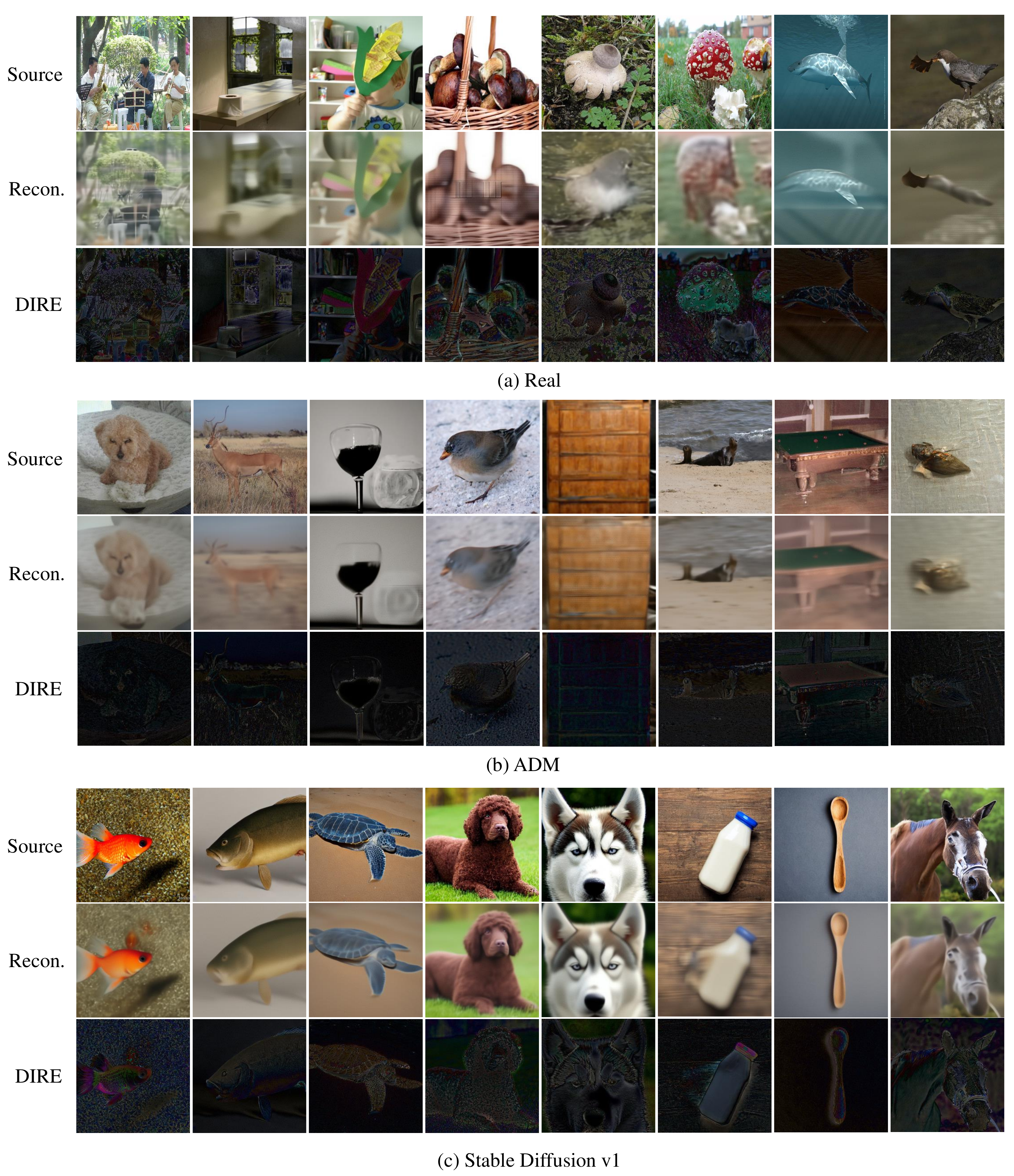}
    \vspace{-1em}
    \caption{The DIRE representation of real images and generated images from ADM and Stable Diffusion v1 with the prompt ``A photo of {class}''~(class from ImageNet~\cite{deng2009imagenet}). For each source image, we visualize its corresponding reconstruction image and DIRE.}
    \label{fig:dire_imagenet}
    \vspace{-5.0mm}
\end{figure*}

We visualize more examples of source images, their reconstructions, and DIREs of real images and generated images from different diffusion models in Figures~\ref{fig:dire_lsun1}, \ref{fig:dire_lsun2}, \ref{fig:dire_lsun3}, \ref{fig:dire_lsun4}, \ref{fig:dire_imagenet}. The
DIREs of real images tend to have larger values compared to diffusion-generated images.

\end{document}